\pdfoutput=1

\documentclass[11pt]{article}

\usepackage[preprint]{acl}
\usepackage{soul}
\usepackage{times}
\usepackage{latexsym}
\usepackage{subfiles}
\usepackage[T1]{fontenc}
\usepackage{multirow} 
\usepackage[utf8]{inputenc}

\usepackage{microtype}

\usepackage{inconsolata}

\usepackage{booktabs}
\usepackage{graphicx}

\usepackage{booktabs}
\usepackage{multirow}
\usepackage{tabularx}
\usepackage{array}
\title{Large Language Models for Document-Level Event-Argument Data Augmentation for Challenging Role Types}

\author{Joseph Gatto, Parker Seemiller, Omar Sharif, Sarah M. Preum \\
  Department of Computer Science, Dartmouth College \\
  \texttt{joseph.m.gatto.gr@dartmouth.edu} \\}

\begin{document}
\maketitle
\begin{abstract}
Event Argument Extraction (EAE) is an extremely difficult information extraction problem --- with significant limitations in few-shot cross-domain (FSCD) settings. A common solution to FSCD modeling is data augmentation. Unfortunately, existing augmentation methods are not well-suited to a variety of real-world EAE contexts including (i) The need to model long documents (10+ sentences) (ii) The need to model zero and few-shot roles (i.e. event roles with little to no training representation). In this work, we introduce two novel LLM-powered data augmentation frameworks for synthesizing extractive document-level EAE samples using \textit{zero in-domain training data}. Our highest performing methods provide a 16-pt increase in F1 score on extraction of zero shot role types. 

To better facilitate analysis of cross-domain EAE, we additionally introduce a new metric, Role-Depth F1 (RDF1), which uses statistical depth to identify roles in the \textit{target domain} which are semantic outliers with respect to roles observed in the \textit{source domain}. Our experiments show that LLM-based augmentation can boost RDF1 performance by up to 11 F1 points compared to baseline methods. 
\end{abstract}

\section{Introduction}

In recent years, there has been considerable progress in the domain of Event Argument Extraction (EAE), as advances in question answering \cite{du-cardie-2020-event}, prompt tuning \cite{ma-etal-2022-prompt}, and semantic graph modeling \cite{yang-etal-2023-amr} have led to state-of-the-art results on EAE benchmarks. However, some simplifying assumptions made in prior work limit their applicability in a real-world setting, including use of sentence-level data, reliance on large-scale annotation, or that training and testing data come from the same domain and thus share a similar label distribution.

Recent works such as \cite{tong-etal-2022-docee, yang-etal-2023-shot} have shown that extracting event arguments from \textit{long documents} given limited training data in a new domain remains extremely challenging. We refer to this problem as few-shot, cross domain (FSCD) document-level EAE (DocEAE) --- where document-level extraction requires identifying event roles across multiple sentences. 

\begin{figure}[!t]
    \centering
    \includegraphics[width=\columnwidth]{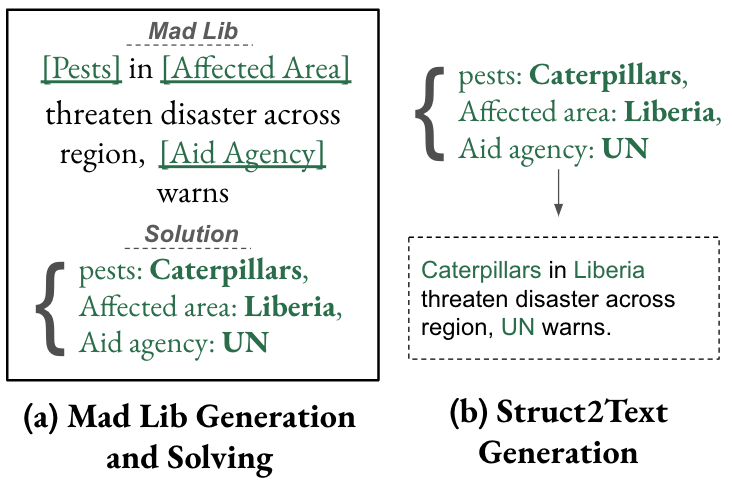}
    \caption{In (a) we leverage LLM's prior knowledge of Mad Libs to generate templated, categorized documents. MadLib solutions can then be used as EAE annotation. (b) We generate event structure data for an event and use LLMs to convert the structure into an EAE document. We employ semantic n-gram matching to align the structure with spans in the generated document. }
    \label{fig:example_madlib}
\end{figure}

A key issue in the FSCD DocEAE problem is the need to model 0-shot roles \cite{yang-etal-2023-shot}, i.e. roles which exist in an event's schema and thus might appear in the test set, but have no representation in the training data. To illustrate this, consider the FSCD DocEE dataset \cite{tong-etal-2022-docee}, where the task is to transfer knowledge from a collection of source domain events (e.g., Road Crash, Celebrity Death, Bank Robbery) to a set of events in a new domain, Natural Disasters (e.g., Droughts, Earthquakes, Tsunamis). Each natural disaster in the FSCD training set only has 5 documents available for training. Across all samples representing the \textit{\textbf{Volcano Eruption}} event, for example, zero documents contain annotation for the role \textit{The State of the Volcano (Dormant or Active)}. This makes this role extremely challenging for many DocEAE models to extract.

One solution to this problem is to strategically generate training samples in order to increase representations of zero and few-shot roles. However, to the best of our knowledge, no prior work has generated DocEAE training samples in a way that can help resolve the aforementioned issues. For example, many related works in EAE data augmentation such as \cite{gao-etal-2022-mask, ma2023star, wang-etal-2023-boosting} only focus on \textit{sentence-level} tasks. The few prior works in DocEAE augmentation do not generate new data samples but rather augment existing samples \cite{LIU2022109904} or use a pre-trained model to weakly label unannotated corpora \cite{liu-etal-2021-machine}.  

In this paper, we address this issue by introducing two strategies for LLM-powered 0-shot DocEAE data augmentation: (i) \textbf{Mad Lib Aug (MLA)} is inspired by the insight that EAE samples can be formulated as the popular game Mad Libs, where the ``blanks" found in Mad Libs correspond to roles in an event ontology. This allows us to leverage LLMs strong conceptual priors of Mad Libs for templated text generation. We visualize the relationship between Mad Libs and EAE in Figure \ref{fig:example_madlib}. We use LLMs to both generate and solve Mad Libs in this framework. (ii) \textbf{Struct2Text (S2T): } is modeled off the synthetic data generation module provided by the LangChain library \footnote{https://www.langchain.com/}, which can transform structured information (e.g. a dictionary of role-argument  mappings) into documents.  Note that both MLA and S2T require no in-domain training samples and can generate data for 0-shot roles given only the event title and role name thus facilitating generalizable cross-domain EAE.

We evaluate each method on DocEE, which to the best of our knowledge is the only FSCD EAE dataset for long documents. Our experimental results show that each approach significantly improves performance on zero-shot roles without sacrificing overall F1 score. We additionally show performance gains on target domain roles which are \textit{semantically different} from those observed in the source domain. To do so, we introduce a new metric, \textbf{Role-Depth F1 (RDF1)}, which uses a statistical depth \cite{seegmiller-preum-2023-statistical} to identify roles in the target domain which are semantic outliers with respect to roles observed in the source domain. For example, the common source-domain role \textit{location} has semantic similarity with other location-based roles, such as \textit{temporary settlement}. Conversely, target roles such as \textit{maximum wind speed} or \textit{Magnitude (Tsunami Heights)} are domain specific and have little semantic overlap with the source training data. We visualize this phenomena in Figure \ref{fig:RDF1-ex}. We argue that RDF1 provides a more useful assessment of cross-domain performance compared to classic F1, which does not capture such semantic nuances. 

Our contributions in this study are as follows: 

\begin{enumerate}
    \item We establish two LLM-powered document-level EAE data augmentation strategies. Our methods can generate novel documents for DocEAE augmentation and are uniquely suited to facilitate generation of zero-shot, cross-domain event roles. Our approach is  generalizable to future low-resource event data, as only the only in-domain information needed for document generation is the event schema. 
    
    \item We show that DocEAE augmentation significantly improves performance on a comprehensive suite of EAE metrics, including a novel evaluation metric, Role-Depth F1 (RDF1). RDF1 characterizes model performance on \textit{semantically outlying} roles. We show that models trained using data generated with MLA and S2T yield significant improvement in RDF1, highlighting the value of augmentation in the cross-domain DocEAE task --- paving the way for future work in this space. 
\end{enumerate} 
\section{Related Work}

\subsection{Document-Level Event Argument Extraction}
Event Argument Extraction (EAE) is a challenging subtask of the Event Extraction (EE) problem, where the goal is to (i) identify  which event(s) occur in the text and (ii) extract event arguments from the text as a structure. Much early work on this problem focuses on sentence level EE, using the popular ACE05 \footnote{https://catalog.ldc.upenn.edu/LDC2006T06} dataset. More recently, some work has focused on document-level datasets such as RAMS \cite{ebner-etal-2020-multi}, WikiEvents \cite{li-etal-2021-document}, and DocEE \cite{tong-etal-2022-docee}. However, RAMS and WikiEvents have significant limitations, as discussed in \cite{tong-etal-2022-docee}. RAMS only uses documents with only 5-sentences and WikiEvents contains annotation for only a limited number of documents. DocEE is the first EAE dataset to contain annotation for a large number of full-length documents, with a mean length of $\approx30$ sentences per document. For this reason, other recent works have chosen to focus their attention on DocEE \cite{yang-etal-2023-shot}. Additionally, DocEE introduces a \textit{cross-domain} split, which is used to facilitate the cross domain EAE analysis in this work (section \ref{experiments}). Hence, we focus on DocEE in this work.

\subsection{Data Augmentation for EAE}
To the best of our knowledge, there is no other work that aims to generate synthetic documents for DocEAE. There are however, adjacent works in the realm of EAE data augmentation. In \cite{liu-etal-2021-machine}, they use pre-trained EAE models to silver-label (i.e. use a pre-trained model to annotate) unannotated documents as additional data augmentation. In \cite{LIU2022109904}, they use pre-trained language models to augment existing samples by masking annotated argument spans and then generating alternate ones. \citet{gao-etal-2022-mask} do the opposite by first masking unannotated spans, then using a pre-trained T5 model to replace such spans as data augmentation. \citet{wang-etal-2023-boosting} perform a reinforcement-learning based solution to sentence-level EE data generation. Contrary to our work, none of the aforementioned methods generate fully novel documents and often rely on augmenting existing samples. \textbf{Works that augment existing samples have limited control of role-type distribution and are not suited to model 0-shot roles, which is a focal point of our work. }

One concurrent work that uses LLMs to generate EAE data for sentence-level tasks is \cite{ma2023star}, where they employ LLMs in a multi-step prompt and self-reflect pipeline. However, there are a number of reasons why we cannot apply this work as a baseline. First, their pipeline leverages information that is unavailable for the DocEE dataset, e.g., event definitions, role definitions, and valid entity types per role. Additionally, they provide the LLM with in-domain demonstrations for trigger generation, when our goal is to use no in-domain data for generation. Finally, their proposed pipeline uses significant LLM-based self-reflection. This is a concern as, for document-level tasks (i) self-reflection makes augmentation significantly more expensive when using proprietary LLMs and (ii) from our experience, self-reflection becomes less effective as document-length increases. Future works may explore how to adapt \cite{ma2023star} to document-level tasks. 

\subsection{Event Extraction Evaluation}

Prior works on EAE in trigger-free datasets use evaluation metrics such as exact match F1 or head noun match F1 \cite{tong-etal-2022-docee}. Recently, there have been some advancements in EAE evalulation for trigger-centric datasets. For example, \cite{huang2023reevaluation} proposed a new metric (ARG-I+) for trigger-based event extraction to better capture whether a role is associated with the correct trigger. In \cite{li2022dual}, they explicitly examine performance on long-tail role types. However, to the best of our knowledge, there are no related works on metrics which can measure an EAE model's performance on semantically outlying roles during cross-domain evaluation. It is crucial to understand to what degree a cross-domain learning algorithm can model truly novel events and not just those which are semantically similar to prior seen events. We propose a novel EAE evaluation metric, Role-Depth F1 (RDF1), to address this gap.

\section{Methods}

In this section we describe both Mad Lib Aug and Struct2Text. Then, we formally introduce our novel metric, Role-Depth F1, which evaluates model performance on semantically outlying role types.

\subsection{Mad Lib Aug (MLA)}

\paragraph{Mad Lib Generation and Slot-Filling: } Recall that Mad Libs are texts interwoven with categorized missing information slots (See Figure \ref{fig:example_madlib} for an example). Thus, to generate an event-driven Mad Lib, we prompt the LLM to write a Mad Lib about a given event type and list each role in the event's schema as a category. Thus for each sample, we aim to generate a document which contains \textit{every possible role} for a given event. Our strategy does not generate documents which intentionally include/exclude certain event role subsets. We choose this strategy for the following reason: A core challenge of document-level EAE data generation is the issue of generating \textit{under-labeled samples}. That is, if the desired length of a document about a single event is large, but the number of roles to be included is small, it is very difficult to prevent the LLM from accidentally generating an unintended role. By providing the LLM with all possible target/desired roles as Mad Lib categories, we mitigate this issue. 

Once we have generated a Mad Lib, we ask the LLM to fill-in the categories (slot-filling) with appropriate information given the document. This is done by prompting the LLM to return a list of \textit{role : argument} pairs which can be slotted into the templated document. We finally do string replacement between the bracketed category names in the Mad Lib, and the LLM-generated solution (shown in Figure \ref{fig:example_madlib}). The result is an event-centric document with annotation for argument extraction. In the event where the LLM generates a Mad Lib with imperfections such as large number of missing arguments or hallucinated roles, we perform a post processing step which is discussed in detail in Appendix \ref{mla_appendix}.

For both Mad Lib Generation and Slot-Filling, we use k=3 documents from the source domain to help the model understand the input/output format. We choose k=3 samples with moderate length and high role-density to minimize API costs for use with proprietary models. Generated data statistics can be found in Table \ref{gen_data_stats}, where we show MLA produces documents with sentence length between 8.2-12.2 across four different LLM backbones.

\begin{figure}
    \centering
    \includegraphics[width=\columnwidth]{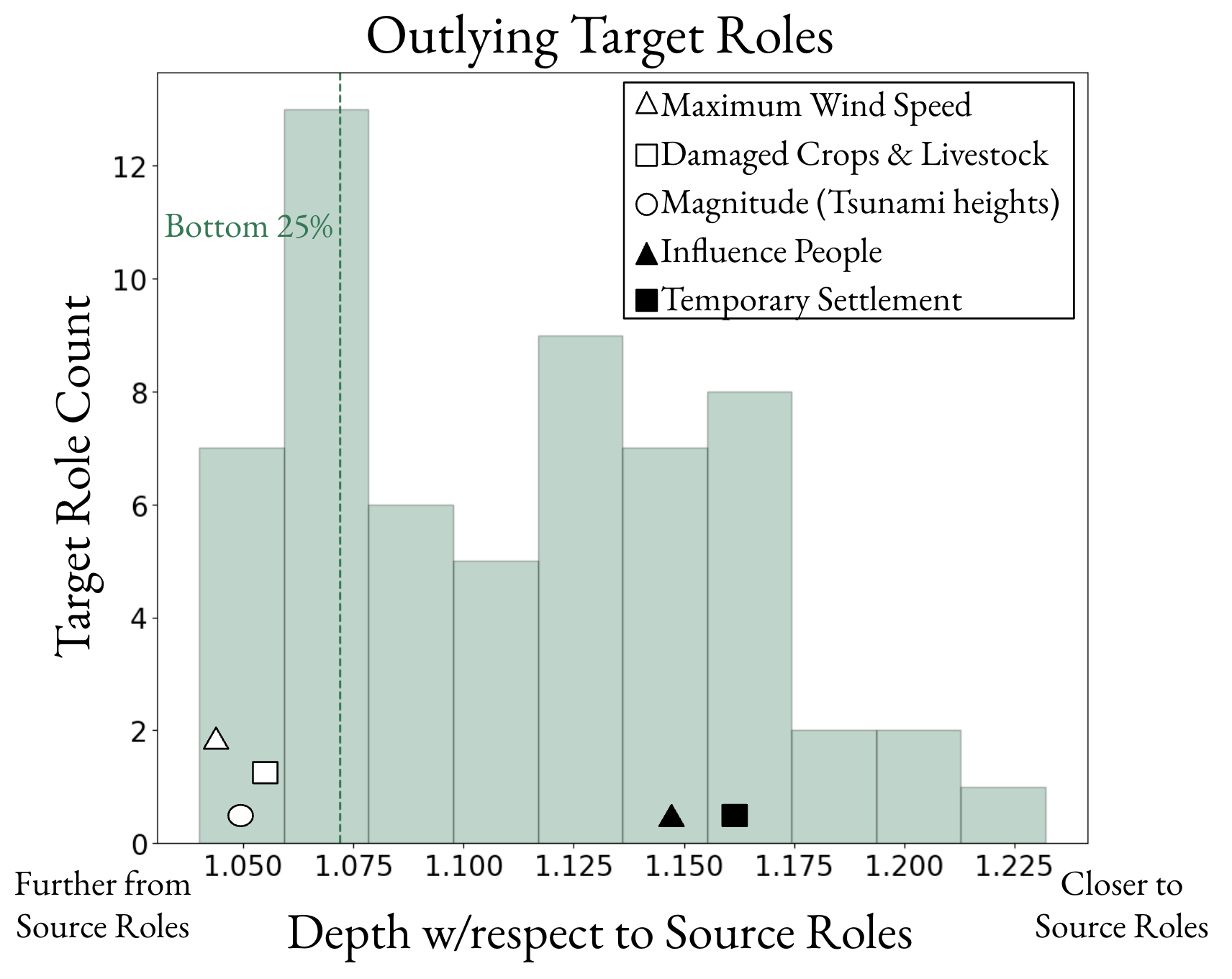}
    \caption{Visualization showing how TTE Depth ranks DocEE roles. We find that roles such as ``Maximum Wind Speed", ``Damaged Crops \& Livestock" and ``Magnitude (Tsunami Heights)" are outliers compared to roles such as ``Influence People" and ``Temporary Settlement". }
    \label{fig:RDF1-ex}
\end{figure}

\begin{table}[!t]
\centering
\resizebox{\columnwidth}{!}{%
\begin{tabular}{@{}cccc@{}}
\toprule
\textbf{Augmentation Method} & \textbf{Sentence Length} & \textbf{Word Count} & \textbf{Avg Roles-Per-Doc} \\ \midrule
MLA-Llama2-7b & 11.9 & 296.8 & 9.1 \\
MLA-Llama2-70b & 12.2 & 304.6 & 10.8 \\
MLA-GPT-3.5 & 8.6 & 188.2 & 9.6 \\
MLA-GPT-4   & 8.2 & 190.9 & 11.7 \\
S2T-GPT-3.5 & 9.8 & 239.1 & 13.6 \\ 
S2T-GPT-4 & 8.8 & 222.1 & 13.4 \\ 
\bottomrule
\end{tabular}%
}
\caption{Generated data statistics from each model. Each column represents the average statistic for each category across all generated documents in the 16x augmentation setting.}
\label{gen_data_stats}
\end{table}
\subsection{Struct2Text (S2T)}

This approach modifies the data augmentation tool provided by LangChain which aims to convert structured data into text. At a high-level, this method uses OpenAI's function calling capability to populate a Python class with event information. We thus create a custom class for each event in the target domain, where the class name is the event name and each role in the event is a class member of type \textit{string}. We closely follow the synthetic data generation strategy implemented by LangChain \footnote{\url{https://python.langchain.com/docs/use_cases/data_generation/}} with a few key modifications (i) We alter the prompt structure to encourage writing of document-level data (i.e. 10-sentence long documents). (ii) We instruct the model to write the documents in the style of a newspaper article to help match the DocEE target domain (iii) We perform semantic n-gram matching to align event structures with generated documents. This is a necessary step as Struct2Text, when used off-the-shelf, does not reliably produce documents with the exact substrings requested when producing document-level data. Our matching algorithm compares the argument in the synthetic event structure with every possible n-gram (up to n=20) in the document and returns the n-gram with the highest cosine similarity using SBERT \footnote{sentence-transformers/all-MiniLM-L6-v2}. We find that for semantic similarity greater than 0.5, this method reliably aligns structure to text. Any arguments which have cosine similarity lower than our threshold are considered to have not been generated and are discarded from the synthetic event record. See Table \ref{gen_data_stats} for S2T generated data statistics.

\subsection{Role-Depth-F1 (RDF1)} \label{RDF1}

RDF1 aims to evaluate how well cross-domain EAE models are predicting role-types which are semantically different from those observed in the source domain. To compute RDF1, we propose the use of transformer-based text embeddings (TTE) depth \cite{seegmiller-preum-2023-statistical} for identifying the degree to which target domain roles deviate semantically from source domain roles. Formally, given a set of source domain roles, $R_S$, and a set of target domain roles, $R_T$, TTE depth assigns a score to each target role $r_t \in R_T$ indicating how well $r_t$ represents $R_S$. 

We embed each role in each set using S-BERT \cite{reimers-gurevych-2019-sentence} to get a set of source role embeddings $S$ and a set of target role embeddings $T$. TTE depth scores each target role embedding $t \in T$ according to

\begin{equation}
\label{eq:depth}
    D_\delta(t,S) := 2 - E_{S}[\delta(t,H)]
\end{equation}

where $H \sim S$ is a random variable with uniform distribution over $S$, and $\delta$ is cosine distance.

TTE depth assigns each target role $r_t \in R_T$ a score, with higher scores indicating that the target role is representative of the source roles $R_S$, and lower scores indicating that the target role is a semantic outlier with respect to the source roles $R_S$.

We consider the 25\% of target domain roles with the lowest TTE depth scores to be "outlying" roles with respect to the source domain roles. We choose 25\% for this TTE depth threshold as it strikes a good balance between filtering for target roles which differ from source domain roles, while maintaining a good amount of roles for evaluation. We explore how the choice of RDF1 threshold affects performance in Table \ref{RDF1_Explore} in the Appendix. By computing F1 only on these roles, i.e. RDF1, we measure the ability of a FSCD EAE model to perform domain transfer to roles which deviate semantically from the source domain. 

In Figure \ref{fig:RDF1-ex}, we visualize how depth scores rank all roles in the target domain with respect to the source domain. Notably, we see that \textit{Maximum Wind Speed}, \textit{Damaged Crops \& Livestock}, and \textit{Magnitude (Tsunami heights)} are the most semantically outlying, with \textit{influence people} and \textit{temporary settlement} being examples of roles closer to source domain roles. This plot justifies the need for RDF1 as a cross-domain evaluation metric, as it follows ones intuition that, for example, \textit{temporary settlement} may have semantic overlap with source domain concepts like \textit{location}. 

In DocEE, 13 roles comprise the set of semantic outliers evaluated in RDF1. 12/13 of these roles have less than 5 training examples and can be considered few-shot roles. We additionally note that the embedding strategy for creating $S$ and $T$ from $R_S$ and $R_T$ is flexible; event ontologies which include additional information such as role descriptions or valid entity types may be used to  enrich the role representations. Since DocEE provides no such information, opting instead for use of semantically-rich role names, the role names themselves are used as $R_S$ and $R_T$.

\begin{table}[!t]
\centering
\resizebox{\columnwidth}{!}{%
\begin{tabular}{@{}cccc@{}}
\toprule
\textbf{Split} & \textbf{\# Samples} & \textbf{\# Events} & \textbf{\# Argument Instances} \\ \midrule
Source Train & 23,630 & 49 & 154,225 \\
\midrule
Target Train & 50    & 10 & 308    \\
Target Dev   & 1,850  & 10 & 11,094  \\
Target Test  & 1,955  & 10 & 13,227  \\ \bottomrule
\end{tabular}%
}
\caption{DocEE data distribution. The source domain contains annotation for over 23k documents across 49 different event types. The target domain contains annotation for documents with 10 distinct event types, all under the common theme of ``Natural Disasters". }
\label{table:data-stats}
\end{table}

\section{Experiments} \label{experiments}
In this section, we describe the details of our experimental setup. Specifically, we introduce the (i) Dataset (ii)  EAE Models, (iii) Baseline Augmentation Strategies, and (iv) Evaluation Metrics.

\subsection{Dataset}
We evaluate our method on the cross-domain split of the DocEE dataset. The DocEE dataset distribution is shown in Table \ref{table:data-stats}. The task is structured to allow pre-training on a large number of source domain events. The source domain in DocEE covers a wide variety of events including Famous Person - Death, Resgination / Dismissal, and Election, amongst others. The goal is to fine-tune this model to a target domain for which there are only very few annotated documents. In this dataset, the target domain is comprised of 10 event types, which are all under the common theme of ``Natural Disasters". This grouping of source and target event domains was chosen to reduce overlap between source and target role types \cite{tong-etal-2022-docee}. The target and source domains have a small 15\% overlap, with common roles such as \textit{date}, \textit{location}, and \textit{economic loss} occurring in both domains. During few-shot cross-domain training, only 5 documents per-event type are provided.

\begin{table*}[ht!]
\centering
\small
\renewcommand{\arraystretch}{1.25}
\begin{tabular}{cl|cccc|cccc}
\toprule
\multicolumn{2}{c|}{} & \multicolumn{4}{c|}{\textbf{Mad Lib Aug (MLA)}} & \multicolumn{4}{c}{\textbf{Struct2Text (S2T)}} \\
\midrule
\textbf{Model} & \textbf{Aug-Backbone} & \textbf{F1} & \textbf{Role F1} & \textbf{Z-Shot F1} & \textbf{RDF1} & \textbf{F1} & \textbf{Role F1} & \textbf{Z-Shot F1} & \textbf{RDF1} \\ 
\midrule
\multirow{7}{*}{BERT-QA}  
                 & \textit{Baseline (No Aug)}   & \textbf{0.31} & 0.24 & 0.19 & 0.20 & 0.31 & 0.24 & 0.19 & 0.20 \\
                 & \textit{Baseline (MTF)}      & \textbf{0.31} & 0.25 & 0.19 & 0.18 & 0.31 & 0.25 & 0.19 & 0.18 \\
                 & \textit{Baseline (Doc-MTF)}  & 0.3 & 0.25 & 0.19 & 0.2 & 0.3 & 0.25 & 0.19 & 0.2 \\
                 \cline{3-10}
                          & \textbf{Llama2-7b}  & 0.29 & 0.27 & $0.28^{\dag}$            & $0.29^{\dag}$                & -    &    - & -             & - \\
                          & \textbf{Llama2-70b} & 0.29 & 0.27 & $0.31^{\dag}$            & $0.29^{\dag}$                & -    &    - & -             & - \\
                          & \textbf{GPT-3.5}    & 0.3  & 0.28 & $0.31^{\dag}$            & $0.29^{\dag}$                & 0.29 & 0.26 & \boldmath{$0.35^{\dag}$} & \textbf{0.28} \\
                          & \textbf{GPT-4}      & 0.3 & \textbf{0.29}  & \boldmath{$0.32^{\dag}$} & \boldmath{$0.31^{\dag}$}     & 0.28 & 0.26 & \boldmath{$0.35^{\dag}$} & \textbf{0.28} \\
\bottomrule
\toprule
\multirow{7}{*}{LF-Seq}   
                & \textit{Baseline (No Aug)}  & 0.22 & 0.06 & 0.0 & 0.01  & 0.22 & 0.06 & 0.0 & 0.01 \\
                & \textit{Baseline (MTF)}     & 0.23 & 0.10 & 0.0 & 0.05  & 0.23 & 0.10 & 0.0 & 0.05 \\
                & \textit{Baseline (Doc-MTF)} & 0.2 & 0.09 & 0.0 & 0.04 & 0.2 & 0.09 & 0.0 & 0.04 \\
                \cline{3-10}
                               & \textbf{Llama2-7b}  & 0.23 & $0.14^{\dag}$          & $0.07^{\dag}$          & $0.12^{\dag}$          & - & - & - & - \\
                               & \textbf{Llama2-70b} & 0.23 & $0.14^{\dag}$          & $0.06^{\dag}$          & $0.12^{\dag}$          & - & - & - & - \\
                               & \textbf{GPT-3.5}    & \textbf{0.24} & $0.14^{\dag}$          & $0.07^{\dag}$          & $0.10^{\dag}$           & 0.23 & \boldmath{$0.15^{\dag}$} & \boldmath{$0.14^{\dag}$} & \boldmath{$0.15^{\dag}$} \\
                               & \textbf{GPT-4}      & 0.22 & \boldmath{$0.15^{\dag}$} & \boldmath{$0.14^{\dag}$} & \boldmath{$0.14^{\dag}$} & 0.22 & \boldmath{$0.15^{\dag}$} & \boldmath{$0.14^{\dag}$} & $0.14^{\dag}$ \\
\bottomrule
\end{tabular}
\caption{Each result is mean of three experimental trials with different random seeds. Bold scores are top performing models. $\dag$ denotes models with statistically significant (p < 0.05) results compared to \textit{all} baselines. Each model is trained with 16x augmentation (n=800 augmented samples). }

\label{tab:results_v2}
\end{table*}

\subsection{EAE Models}

In this study, we evaluate the impact of EAE data augmentation using the following two EAE models motivated by the DocEE Paper \cite{tong-etal-2022-docee}. 

\begin{itemize}
    \item \textbf{LongFormer(LF)-Seq:} This model treats EAE as a token labeling task. Specifically, for a document $D$ with tokens $\{t_0, \dots, t_n\} \in D$, we label argument spans using the label space $\{r_0,  \dots, r_n\} \in R$, where a given $r_i$ denotes an event role. Given that we are processing long documents, we use the LongFormer-base (149M parameters) \cite{Beltagy2020Longformer} architecture, which can process up to 4096 tokens. We use this model as it has been shown to out-perform BERT-based token classifiers on DocEAE \cite{tong-etal-2022-docee}.
    
    \item \textbf{BERT-QA: } This model treats EAE as an extractive question-answering task, inspired by  \cite{du-cardie-2020-event}. In this formulation, the question is the argument role, and the context is the document. The goal is to predict the span(s) $(start_i, end_i)$ corresponding to the argument of the role in the document. Note that for this baseline, we use a BERT-base (110M parameters) \cite{devlin-etal-2019-bert} model as in the original DocEE paper. 
\end{itemize}

We note that the use of two baseline methods is in-line with prior works on data augmentation for Event Extraction \cite{gao-etal-2022-mask}. Additionally, we highlight that DocEE is a \textit{trigger-free} event dataset, meaning that the document is itself event-centric and not grounded to any one trigger mention. This restricts our capacity to evaluate on trigger-dependent  EAE baselines \cite{parekh-etal-2023-geneva,huang2023reevaluation}. 

For each experiment, we first fine-tune the model on the source-domain training set, then on the few-shot target dataset. Target domain models are chosen using validation set performance. We train all models on Google Colab using an NVIDIA A100 GPU. Throughout the course of the project we estimate an upper-limit of 100 compute hours. Additional information regarding model hyperparameters, compute resources, and training details can be found in Appendix \ref{baseline_models_apendix}.

\subsection{Baseline Augmentation Methods}
While this work is the first to explore document-level synthetic data generation, we compare our approach to a relevant EAE augmentation method developed for a sentence-level task and can be adapted for DocEE: \textbf{Mask-then-Fill (MTF)} \cite{gao-etal-2022-mask}. MTF augments DocEE documents by masking a single contiguous span of non-annotated text in each document and infilling the mask using a T5 \cite{2020t5} model fine-tuned on the Gigaword corpus \cite{graff2003english}. Since MTF was developed for sentence-level tasks, we additionally implement a document-level variant: Doc-MTF. \textbf{Doc-MTF} is similar to MTF, except the T5 model is trained to fill masks of larger contiguous spans of non-annotated text in documents from the MultiNews dataset \cite{alex2019multinews}, which is more similar in style and length to DocEE texts than the original Gigaword corpus. Additionally, Doc-MTF is trained to fill multiple masks in a single document, which enables more diverse augmentations than the original single-mask variant.

\subsection{LLMs for Data Generation}

Since MLA uses an in-context learning-based prompting strategy, we can leverage both small open-source LLMs as well as high-powered proprietary LLMs. We thus evaluate MLA in the context of (i) Llama2-7b, (ii) Llama2-70b (iii) GPT-3.5 (iv) GPT-4. S2T, on the contrary, relies on function calling techniques which OpenAI models are specifically fine-tuned to utilize, making it challenging to integrate with smaller open-source LLMs. We thus only investigate S2T in the context of GPT-3.5 and GPT-4. We use temperature = 1 for each LLM experiment. Please refer to Appendix \ref{augmentation_appendix} for additional details on LLMs used in this study.

\subsection{Evaluation Metrics}

We evaluate each model using an exact match F1 score as motivated by prior work \cite{tong-etal-2022-docee}. Additionally, we evaluate on a diverse suite of EAE metrics which aim to capture performance on challenging role types. Inspired by \cite{li2022dual}, for each of the following metrics, we subset evaluation data based on role-type and compute mean F1 score per-role. 

\begin{itemize}
    \item \textbf{Role-F1}: We compute the F1 score of each role type, then report the macro average across all roles. This metric removes bias towards prediction of common roles. Role bias is removed in this setting as high-frequency roles no longer disproportionately affect overall F1 score.  There are 60 total roles in the target domain. 
    \item \textbf{0-Shot F1}: This metric identifies performance on roles with no representation in the DocEE target training set. We thus compute F1 score of each role with 0 target domain training samples, then report macro average. There are 9 roles in this subset. 

    \item \textbf{RDF1}: Our novel metric, RDF1, computes the F1 score of each role in the bucket of semantic outliers, then reports the macro average. Statistical depth identifies 13 roles as outliers which are used in this evaluation. See section \ref{RDF1} for additional details. Note that 92\% of these roles are few-shot, meaning they have less than 5 training examples. 
    
\end{itemize}

For each metric, we consider the exact match between a \textit{set} of predicted and ground truth tuples consisting of (document id, role type, argument). We report the mean of three experimental trails with different random seeds for each metric. We perform statistical significance testing by using the Wilcoxon signed rank test. For each experiment, we concatenate the F1 scores for each role/document across each experiment trial and use this to test statistical significance. We place a $\dag$ marker next to results in Table \ref{tab:results_v2} which have statistically significant differences (p < 0.05) in distributions between all baseline scores.

\section{Results}

\paragraph{Significant Improvement in F1 Score on Zero-Shot Roles: } We find that both MLA and S2T have the capacity to significantly improve the breadth of roles which can be accurately predicted by a DocEAE model. On BERT-QA, MLA and S2T provide a 13-pt and 14-pt boost in Z-Shot F1 score respectively. The LF-Seq model also achieves a 14-pt boost in F1 score for MLA and S2T when compared to baseline models. We note that there is little difference between the results using GPT-3.5 and Llama2-70b across all experiments, highlighting the capacity of MLA to work with open-source LLMs.  In general, we find that BERT-QA out-performs LF-Seq in all settings, which is intuitive as BERT-QA is only focused on one role per-inference, as where LF-Seq has the potential to accidentally assign labels from other event types.

\paragraph{Data Augmentation Improves Prediction of Semantically Outlying Roles: } Both MLA and S2T significantly improve performance on semantically outlying RDF1 roles. Notably, our smallest LLM, Llama2-7b, which has less representational capacity compared to other LLMs still generates samples with highly useful training signal. MLA and S2T improve performance over the nearest baseline on RDF1 by 11-pts and 8-pts respectively on the BERT-QA model. On the LF-Seq model, we see a 9-pt and 10-pt boost in RDF1 from MLA and S2T respectively. This result confirms that LLM-based DocEAE generation significantly enhances cross-domain learning. We note that for all metrics, MLA and S2T models achieve similar F1 scores, highlighting how this method leads to more evenly distributed role performance vs baseline models which are skewed to overall F1.

\paragraph{Trade-Off Between Overall F1 and Role-Specific Metrics: } Our results show that across all experiments there is no statistically significant difference in overall F1 score between baseline and augmented models. We find that roles which have significant representation in both the training and testing set often do not benefit from augmentation. This is intuitive, as the model can overfit to high training frequency roles and improve overall F1 since \textit{the test set is dominated by a small set of common roles}. For example, the role \textit{Date}, which appears 28 times in the training set (3rd most frequently occurring training role), and 1841 times in the test set, does not benefit from augmentation. A model can thus overfit to Date and gain significant performance increase in overall F1 score as this role type dominates the test set. Additionally, it may be hard to generate diverse augmentations for date, as DocEE considers arguments such as ``Now" or ``just as the lunch hour was ending Tuesday" to be dates, while MLA GPT-4 exclusively generates things in some variation of the Day-Month-Year format. This is in contrast to the role \textit{The State of the Volcano (Dormant or Active)} which occurs only 52 times in the test set, and never occurs in the training set. In this situation, the LLM significantly improves model performance as it has no prior exposure to such a role-type. Thus, if it is more important to the end user of a DocEAE system that a wide variety of roles can be accurately extracted, both MLA and S2T are well-suited to improve performance in low-resource settings. 

\paragraph{Performance Plateaus at 8x Augmentation: } In Table \ref{scaling}, we explore how scaling up augmentation impacts performance on three different metrics. We find that increasing the scale from 0x to 4x provides significant improvement. Additionally, scaling from 4x to 8x improves performance on Z-Shot F1 and RDF1. However, further scaling the data to 16x has limited impact and can even reduce performance on Z-Shot F1 and RDF1. We believe the limits of scaling may be driven by the 0-shot nature of the generation, where it can be challenging for an LLM to come up with diverse arguments given only event type and role name information. For example, many of the events pertaining to natural disasters have the role of ``Aid Agency", for which GPT-4 overwhelmingly generates ``Red Cross" or ``American Red Cross" as the argument, as it is one of the most popular Aid Agencies in the world. Future works may address this gap via external knowledge injection during the data generation process.

\begin{table}[!t]
\centering

\label{my-label}
\begin{tabular}{@{}ccccc@{}} 
\toprule
Aug   & Scale & F1 & Z-Shot F1 & RDF1 \\ \midrule

Baseline & - & 0.31 & 0.19 & 0.20 \\
\midrule
\multirow{3}{*}{MLA-GPT4}& 4x   & 0.31      & 0.30      & 0.31    \\
                         & 8x   & 0.31      & 0.32      & 0.32    \\
                         & 16x  & 0.30      & 0.32      & 0.31    \\
                         \midrule 
\multirow{3}{*}{S2T-GPT4}& 4x   & 0.30      & 0.33      & 0.29    \\
                         & 8x   & 0.30      & 0.36      & 0.31    \\
                         & 16x  & 0.28      & 0.35      & 0.28    \\ \bottomrule
\end{tabular}
\caption{Exploring how augmentation scale affects performance. We find that for Z-Shot F1 and RDF1, 8x augmentation provides the strongest performance for both models. There are thus limits to how much scaling augmentation can improve performance. }
\label{scaling}
\end{table}
\section{Conclusion}

In this study, we showcase the capacity of LLMs to generate event-centric document-level argument extraction data. Our LLM-driven framework produces high-quality DocEAE annotation and significantly improves performance on a variety of cross-domain DocEAE metrics. We find that MadLib Aug is a promising method of generating categorized, templated documents. This method is shown to be successful using both open-source and proprietary models, making it an accessible form of DocEAE augmentation. S2T, on average, provides slightly higher scores than MLA on various metrics and is thus a powerful method of DocEAE data generation. In the future, we hope to explore how Mad Lib Aug can generate documents with greater diversity, from both the Mad Lib Generation perspective (e.g. generating narrative works) and the Mad Lib Slot-Filling perspective (e.g. the Mad Lib solver takes the perspective of a specific person). We also wish to explore adding role definitions to the generation process for both MLA and S2T. Additional future work may aim to extend this method to work with trigger-centric event datasets.

\section{Limitations}

This work is limited in that we do not explore the capacity of our model to generate large-scale data augmentation with fine-grained control over the role distribution. This was deemed out of scope for this work due to the cost of data generation and the focus on a few-shot task. Additionally, this is an extremely difficult sub-task which requires a more targeted solution. Future works will explore scaling our framework in this way, on both in and cross-domain formatted tasks. 

A limitation of the S2T model is it's reliance on the OpenAI function calling capabilities. Future works may wish to explore how to adapt this method to an open-source LLM which has been fine-tuned for function calling in a manner that facilitates DocEAE data generation. Another  general limitation of this study is that we do not thoroughly explore the capacity of these methods to generate multi-argument roles. For example, if there are two different causes to an event, having the MLA model generate templates for [Cause \#1] and [Cause \#2]. Such extensions  will be the focus of future work. 

Finally, this work is limited in that it does not fully explore the capacity of LLMs to perform related types of augmentation such as style transfer (e.g. generating Mad Libs in a variety of styles). This may improve performance even further and will be a subject of future work.


\bibliography{anthology, custom}
\appendix
\appendix

\section{Baseline Models} \label{baseline_models_apendix}
Details of our baseline models are found in the following sections. 

\subsection{BERT-QA}
All of the BERT-QA experiments are run using the Huggingface \cite{huggingface} Transformers library. We use a BERT-base model (110m parameters) \footnote{bert-base-uncased} as our base model. Given that BERT has a max token length of 512, we process longer documents in chunks using a stride of 50 tokens. We then fine-tune the model for 3 epochs on the source domain using default Huggingface Trainer parameters \footnote{\url{https://huggingface.co/docs/transformers/main_classes/trainer}}. During cross-domain fine-tuning, we again train for 3 epochs on target domain data with default parameters and use the model with the best validation loss score for test set prediction. This typically converges after 1 epoch. To help encourage extraction of multi-argument roles, we only include unique arguments as training examples for QA-based extraction. During inference, we take up to 2 predictions per-question, as some documents express more than one argument per role. We perform inference using the Huggingface question-answering pipeline. We consider roles present in test set during evaluation. We use a batch size of 48 for all BERT-QA experiments and train our models on Google Colab use an NVIDIA A100 GPU.

\subsection{LongFormer-Seq}
All LongFormer-Seq experiments use the LongFormer-base (149M parameters) \footnote{allenai/longformer-base-4096} and are run using the Huggingface Transformers library. As with BERT-QA, we fine-tune the model for 3-epochs on the source training set with default parameters. During cross-domain fine-tuning, we fine-tune for a maximum of 10 epochs using validation F1 score from SeqEval \cite{seqeval} to determine the best model. Note we train this model longer as a new classification head must be learned for cross-domain learning. We use a batch size of 4 for all LongFormer experiments and train our models on Google Collab use an NVIDIA A100 GPU. 

\subsection{Evaluation Metrics}
We use an implementation of F1 score for EAE from the OmniEvent library \cite{peng-etal-2023-devil}. When computing F1, since the same argument can be expressed in slightly different ways throughout a document, we normalize the argument string before evaluation using a function provided by the DocEE authors. This makes outputs lowercase, removes articles, white space, and punctuation. For example, If the prediction was (0, date, The 14th of May.), normalization would return (0, date, 14th of may). This is helpful since the same argument can be expressed in many slightly different ways throughout a long document.

\section{Data Augmentation Methods} \label{augmentation_appendix}
In this section, we discuss various hyperparameters and design choices of our two generative data augmentation models, Mad Lib Aug and Struct2Text. 

\subsection{Hyperparameters}
For each model from OpenAI, we use temperature=1 with all other parameters left default. We use k=3 in-context learning examples when applicable. For open-source Llama models we restrict generation to 512 tokens. Future works may explore generation of even longer documents. 

\subsection{Mad Lib Aug Post Processing} \label{mla_appendix}
To ensure high-quality data, we post-process our generated Mad Libs. An error that can occur during this process is \textit{hallucinated role types}. For example, if the LLM invents a new role not found in the event schema when generating the Mad Lib, we may handle this in two ways: (i) we can use SBERT \cite{reimers-gurevych-2019-sentence}, a popular textual embedding method, to encode the hallucinated category name and compute the cosine similarity between it and all possible role names for a given sample. We can often reliably match the generated category to the true category using this method for mild hallucinations (e.g. mapping ``The place" back to ``Location"). (ii) For smaller LLMs we find hallucinations can be more extreme, thus we simply remove the sentence containing the hallucinated role from the document. Since DocEE is largely comprised of news articles, simply removing the sentence is a non-destructive way to deal with hallucination as the documents are often comprised of a sequence of factual statements. In our experiments, we focus on method (ii) as it is less ambiguous and works with all models. 

Additionally, the generation of a longer text will be particularly sensitive to the randomly drawn few-shot sample selection. Thus, there can be outlying generation attempts where Mad Libs are generated (i) with little to no role information (ii) with incorrectly used brackets (iii) with improperly formatted solutions. When this happens, we simply draw a new set of few-shot examples and try again. This issue is predominantly found when using Llama2-7b. For larger models such as GPT-4, this almost never happens. We restrict the model to a maximum of 5 retries during such situations. 

\subsection{Struct2Text Post Processing} \label{mla_appendix}
Similar to MLA, S2T can suffer from random generation errors. Unlike MLA, S2T's errors are largely constrained to JSON parsing issues which occurring during use of function calling. We thus similarly allow for 5 attempts at sample generation, which is shown to be enough to handle random JSON errors. 

\subsubsection{Prompts}
\paragraph{Mad Lib Generation: }

\begin{quote}
Write a madlib for a ``\{event\_type\}" event. Use the following categories for Mad Lib blanks. Make sure all categories appear in the MadLib. Do not generate any other categories.:

Categories: \{categories\} 

Madlib:
\end{quote}

\paragraph{Mad Lib Category In-Filling}
\begin{quote}
Solve the MadLib. Fill-in missing information in the document.
Return your answer as a bulleted list of the format  ``\#\#\# CATEGORY: ANSWER". 

Make sure the following categories are in your solution: \{categories\}

Madlib: \{document\}

MadLib Solution:
\end{quote}

\section{Exploring The Effect of RDF1 Percentage }

To motivate our choice of 25\% as the outlying role threshold for RDF1, we recompute the results using 4 separate thresholds in the range of $
\pm 10\%$. We find that for the DocEE dataset, the threshold does not significantly alter performance. 

\begin{table}[!h]
\centering
\small
\begin{tabular}{llllll}
\toprule

15\% & 20\% & 25\% & 30\% & 35\% & Aug Method \\
\midrule
 0.24 & 0.22 & 0.2 & 0.19 & 0.2 & Baseline \\
 0.24 & 0.2 & 0.18 & 0.18 & 0.2 & MTF \\
 0.27 & 0.27 & 0.28 & 0.26 & 0.26 & S2T \\
 0.3 & 0.3 & 0.31 & 0.31 & 0.3 & MLA \\
\bottomrule
\end{tabular}
\caption{RDF1 scores at varying thresholds for outlying role designation. We find that models are relatively stable given slight fluctuations in RDF1 threshold. }
\label{RDF1_Explore}
\end{table}

\section{Example Data} \label{example_data_section}

\begin{figure*}[!h]
    \centering
    \includegraphics[width=\textwidth]{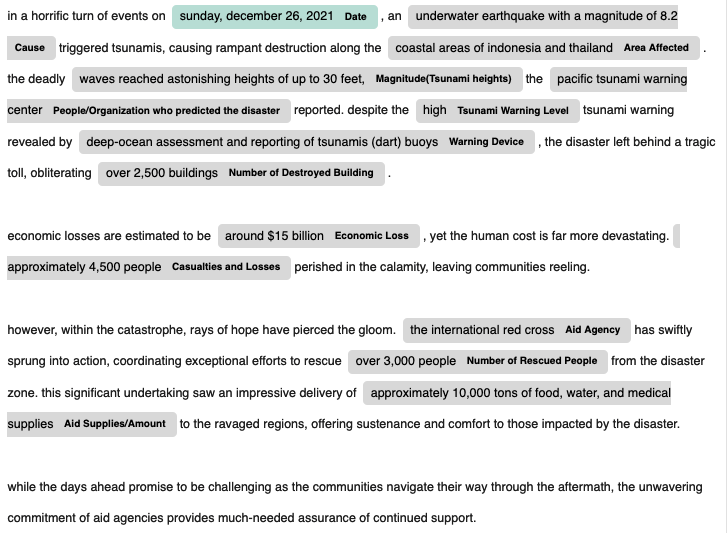}
    \caption{Example data for Tsunami Event using Struct-2-Text (GPT-4). }
    \label{fig:enter-label}
\end{figure*}

\begin{figure*}[!h]
    \centering
    \includegraphics[width=\textwidth]{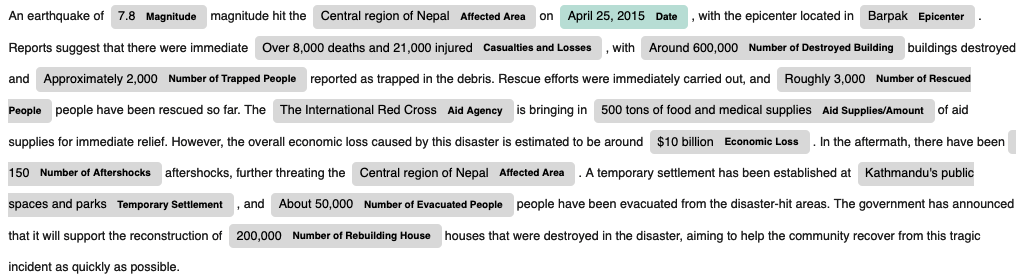}
    \caption{Example data for Earthquake event using Mad Lib Aug (GPT-4). }
    \label{fig:enter-label}
\end{figure*}

\begin{figure*}[!h]
    \centering
    \includegraphics[width=\textwidth]{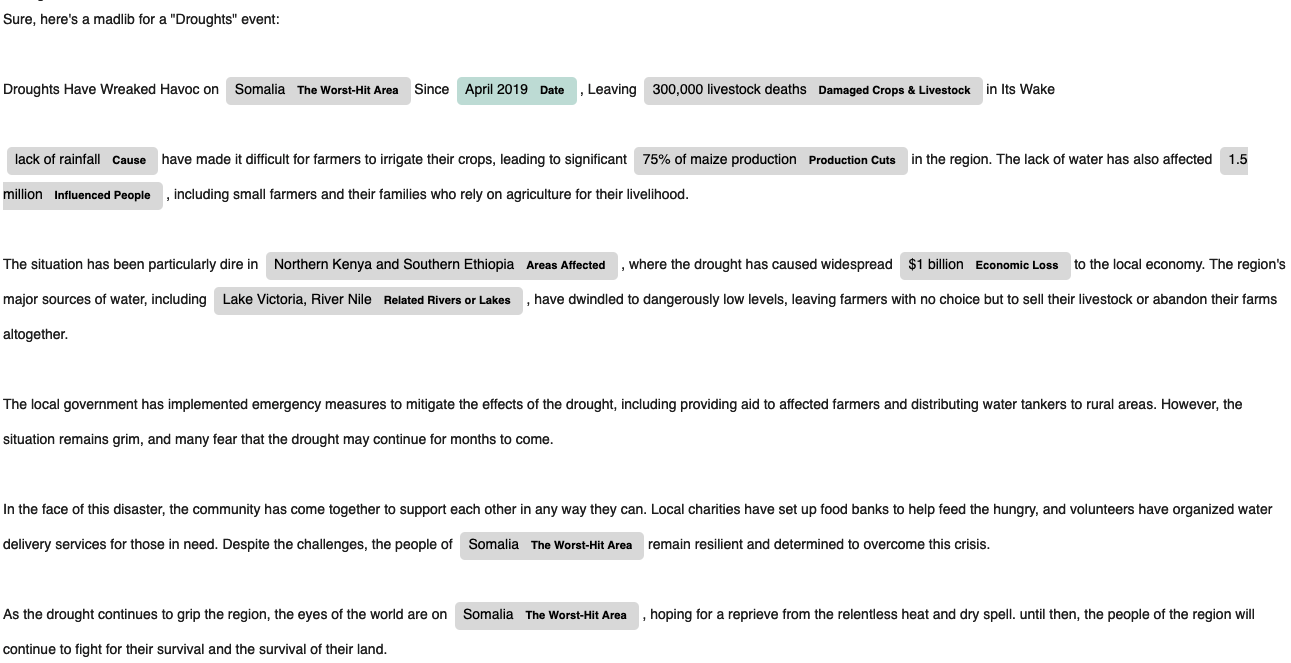}
    \caption{Example data for Droughts Event using Mad Lib Aug (Llama-7b). Note that sample quality suffers for smaller LLMs. However, such samples still serve as useful training signal as shown in our results.}
    \label{fig:enter-label}
\end{figure*}

\begin{figure*}[!h]
    \centering
    \includegraphics[width=\textwidth]{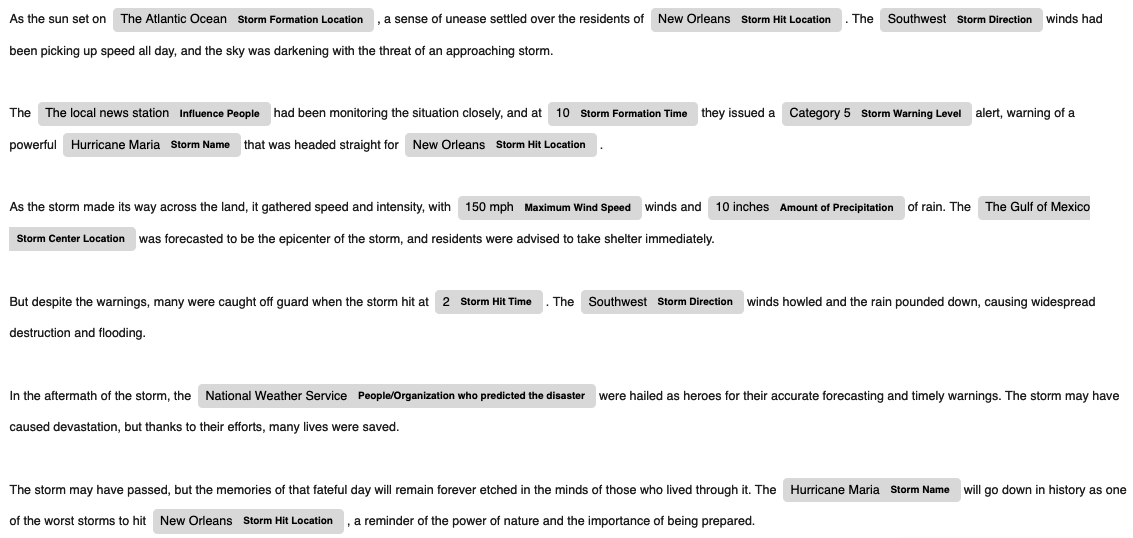}
    \caption{Example data for Storm Event using Mad Lib Aug (Llama-70b). }
    \label{fig:enter-label}
\end{figure*}

\end{document}